\pdfoutput=1

\documentclass[11pt]{article}

\usepackage[]{acl}

\usepackage{times}
\usepackage{latexsym}

\usepackage[T1]{fontenc}

\usepackage[utf8]{inputenc}

\usepackage{microtype}

\usepackage{inconsolata}

\newcommand{\method}{ContraSim}
\usepackage{amsmath}
\usepackage{amssymb}
\usepackage{mathtools}
\usepackage{amsthm}
\usepackage{subcaption}
\usepackage{booktabs} 
\usepackage{bm}
\usepackage{multirow}
\usepackage{xcolor}
\DeclareMathOperator*{\argmax}{arg\,max}

\title{\method{} -- Analyzing Neural Representations Based on Contrastive Learning}

\author{Adir Rahamim
\hspace{10em}
  Yonatan Belinkov \\
    \texttt{adir.rahamim@cs.technion.ac.il} \hspace{2em} 
  \texttt{belinkov@technion.ac.il} \\
  Technion -- Israel Institute of Technology}

\begin{document}
\maketitle
\begin{abstract}
Recent work has compared neural network representations via similarity-based analyses to improve model interpretation.
  The quality of a similarity measure is typically evaluated by its success in assigning a high score to representations that are expected to be matched. 
  However, existing similarity measures perform mediocrely on standard benchmarks. 
  In this work, we develop a new similarity measure, dubbed \method{}, based on contrastive learning. In contrast to common closed-form similarity measures, \method{} learns a parameterized measure by using both similar and dissimilar examples. 
 We perform an extensive experimental evaluation of our method, with both language and vision models, on the standard layer prediction benchmark and two new benchmarks that we introduce: the multilingual benchmark and the image--caption benchmark. In all cases, \method{} achieves much higher accuracy than previous similarity measures, even when presented with challenging examples. Finally, \method{} is more suitable for the analysis of neural networks, revealing new insights not captured by previous measures.\footnote{Code is available at: \url{https://github.com/technion-cs-nlp/ContraSim}.}
\end{abstract}

\section{Introduction}

Representation learning has been key in the success of deep neural networks (NNs) on many tasks.
However, the resulting representations are opaque and not easily understood. 
A recent line of work analyzes internal representations by comparing two sets of representations, for instance from two different models. 
The choice of similarity measure is crucial and much work has been devoted to developing various such measures  \cite{raghu2017svcca,morcos2018insights,kornblith2019similarity,wu2020similarity}. Similarity-based analyses may shed light on how different datasets, architectures, etc., change the model's learned representations, and improve their interpretability. For example, a similarity analysis showed that lower layers in different language models are more similar to each other, while fine-tuning affects mostly the top layers \citep{wu2020similarity}. 

Various similarity measures have been proposed for comparing representations, among them the most popular ones are based on centered kernel alignment (CKA) \citep{kornblith2019similarity} and canonical correlation analysis (CCA)  \citep{Hotelling1936RelationsBT, morcos2018insights}. They all share a similar methodology: given a pair of feature representations \emph{of the same input}, they estimate the similarity between them, without considering other examples. However, they all perform mediocrely on standard benchmarks. Thus, we might question the reliability of their similarity scores, as well as the validity of interpretability insights derived from them. 

Motivated by this, we introduce \method{}, a new similarity measure for interpreting NNs, based on contrastive learning (CL) \citep{chen2020simple, he2020momentum}. Contrary to prior work \cite[e.g.,][]{raghu2017svcca,kornblith2019similarity}, which defines closed-form general-purpose similarity measures, \method{} is a learnable similarity measure that uses examples with a high similarity (the \textit{positive} set) and examples that have a low similarity (the \textit{negative} set), to train an encoder that maps representations to the space where similarity is measured. In the projected space, representation similarity is maximized with positive examples and minimized with negative examples. Our approach allows specializing the similarity measure to a particular domain, to obtain a more reliable and specific analysis. The similarity between projected representations is determined using a simpler closed-form measure.

We experimentally evaluate \method{} on standard benchmark for similarity measures -- the layer prediction benchmark \cite{kornblith2019similarity}, and two new benchmarks we introduce in this paper: the multilingual benchmark and the image--caption benchmark. In experiments with both language and vision models and multiple datasets, \method{} outperforms common similarity measures. In addition, we investigate a more challenging scenario, where during evaluation instead of choosing a random sentence, we retrieve a highly similar sentences as confusing examples, using the Facebook AI Similarity Search (FAISS) library \cite{johnson2019billion}. While other similarity measures are highly affected by this change, our method maintains a high accuracy with a very small degradation. We attribute this to the highly separable representations that our method learns. Even when \method{} is trained on data from one domain/task and evaluated on data from another domain/task, it achieves superior performance.

Through ablations, we demonstrate that the CL procedure is crucial to the success of the method and only maximizing the similarity of positive examples is not sufficient. Furthermore, we demonstrate that \method{} reveals new insights not captured by previous similarity measures.
For instance,  CKA suggests that the BERT  language model \cite{devlin2018bert} is more similar to vision models than  GPT-2 \cite{radford2019language}. Our analysis indicates that both BERT and GPT-2 create representations that are equally similar to the vision ones.  



\section{Related Work} \label{sec:related-work}

 Many studies have used similarity measures for the interpretability of NNs.
 For instance, \citet{kornblith2019similarity} showed that adding too many layers to a convolutional neural network, trained for image classification, hurts its performance. Using CKA, they found that more than half of the network's layers are very similar to the last. 
They further found that two models trained on different image datasets (CIFAR-10 and CIFAR-100, \citealt{krizhevsky2009learning})  learn representations that are similar in the shallow layers. Similar findings were noted for language models by \citet{wu2020similarity}. 
The latter also evaluated the effect of fine-tuning on language models and found that the top layers  are most affected by fine-tuning.
\citet{kornblith2019similarity} and \citet{morcos2018insights}  found that increasing the model's layer width results in more similar representations between models.
\citet{raghu2017svcca} provided an interpretation of the learning process by examining how similar representations were during the training process compared to final representations. They found that networks converge from bottom to top.
In all this line of work, the similarity is computed using only similar examples, using functional closed-form measures. In contrast, we use both positive and negative samples in a learnable similarity measure, which allows adaptation to specific tasks.

Separate work employs contrastive learning for representation learning \citep{he2020momentum, chen2020simple} and data retrieval \cite{karpathy2014deep, huang2008active}. In contrast to that line of work, we borrow the contrastive learning formulation for similarity-based interpretations of deep networks.


\section{Problem Setup}
Let $\mathbb{X}=\{(x_1^{(i)}, x_2^{(i)}) \}_{i=1}^N$ denote a set of $N$ examples, and  $\mathbb{A}=\{(\bm{a}_1^{(i)}, \bm{a}_2^{(i)}) \}_{i=1}^N$ the set of representations generated for the examples in $\mathbb{X}$. 
A representation is a high-dimensional vector of neuron activations. Representations may be created by different models, different layers of the same model, etc.
For instance,  $x_1^{(i)}$ and  $x_2^{(i)}$ may be the same input, with $\bm{a}_1^{(i)}$ and $\bm{a}_2^{(i)}$ representations of that input in different layers. 

Our goal is to obtain a scalar similarity score, which represents the similarity between the two sets of representations, $\bm{a}_1^{(i)}$ and $\bm{a}_2^{(i)}$, and ranges from $0$ (no similarity) to $1$ (identical representations). 
That is, we define $\bm{X}_1 \in \mathbb{R}^{N \times p_1}$ as a matrix of $p_1$-dim activations of $N$ data points, and $\bm{X}_2 \in \mathbb{R}^{N \times p_2}$ as another matrix of $p_2$-dim activations of $N$ data points. We seek a similarity measure, $s(\bm{X}_1, \bm{X}_2)$. 
\section{\method{}}
\label{sec:method}
In this section we introduce \method{}, a similarity index for measuring the similarity of neural network representations. Our method uses a trainable encoder, which first maps representations to a new space and then measures the similarity of the projected representations. Formally, let $e_\theta$ denote an encoder with trainable parameters $\theta$, and assume two representations $\bm{a}_1$ and $\bm{a}_2$. 
In order to obtain a similarity score between $0$ and $1$,
we first apply L2 normalization to the encoder outputs: $\bm{z}_1 = e_\theta(\bm{a}_1) / \| e_\theta(\bm{a}_1) \|$ (and similarly for $\bm{a}_2$). 
Then their similarity is calculated as: 
\begin{equation}
\label{sim}
    s(\bm{z}_1, \bm{z}_2)
\end{equation}
where $s$ is a simple closed-form similarity measure for two vectors. Throughout this work we use dot product for $s$.   

For efficiency reasons, we calculate the similarity between batches of the normalized encoder representations, dividing by the batch size $n$:   
\begin{equation}
    \frac{1}{n} \sum_{i=1}^n \left(\bm{z}_1^{i} \cdot \bm{z}_2^{i} \right)
\end{equation}

If the representations $a_1$ and $a_2$ have the same dimensionality, \method{} can be trained with a single encoder shared for both representations. In the case that $a_1$ and $a_2$ have different dimensions, two different encoders are trained, $e_{\theta1}$ and $e_{\theta2}$, one for each representation set. By that, \method{} can calculate the similarity of representations with different dimensionality, as each encoder has a different input dimension, but both encoders share the same output dimension. Experiments with representations with different dimensionality are in Section \ref{different_dim}.

\paragraph{Training.} 
None of the similarity measures commonly used in NNs analysis uses negative examples to estimate the similarity of a given pair (Section~\ref{sec:related-work}). Given two examples, these measures output a scalar that represents the similarity between them, without leveraging data from other examples. However, based on knowledge from other examples, we can construct a better similarity measure. 
In particular, for a given example $x^{(i)} \in \mathbb{X}$ with its encoded representation $\bm{z}_i$, we construct a set of \textit{positive} example indices, $P(i) = \{p_1, ..., p_q\}$, and a set of \textit{negative} example indices, $N(i) = \{n_1, ..., n_t\}$. 
The choice of these sets is task-specific and allows one to add inductive bias to the training process. 

We train the encoder to maximize the similarity of $\bm{z}_i$ with all the positive examples, while at the same time making it dis-similar from the negative examples. We leverage ideas from contrastive learning \citep{chen2020simple, he2020momentum}, and minimize the following objective:
\begin{equation}
\mathcal{L} =  \sum_{i\in I} \frac{-1}{|P(i)|}  \log \frac{\sum_{p\in P(i)} \exp (\bm{z}_i \cdot \bm{z}_p/ \tau)}{\sum_{n\in N(i)} \exp (\bm{z}_i \cdot \bm{z}_n/ \tau)}
\label{cl_loss}
\end{equation}
with scalar temperature parameter $\tau > 0$. Here $\bm{z}_p$ and $\bm{z}_n$ are normalized encoder outputs of representations from the positive and negative groups, respectively. It is important to notice that we do not alter the original model representations, and the only trainable parameters are the encoder $e$ parameters, $\theta$. 
Our work uses negative examples and a trainable encoder for constructing a similarity measure. We evaluate these two aspects in the experimental section and show that using negative examples is an important aspect of our method. Combining the two leads to a similarity measure that outperforms current measures.

\section{Similarity Measure Evaluation}
\label{eval}
For evaluation, we use the known layer prediction benchmark and two new benchmarks we design: the multilingual benchmark and the image--caption benchmark. We further propose a strengthened version of the last two using the FAISS software.

\subsection{Layer prediction benchmark}
\begin{figure}[t]
\centering
\includegraphics[width=0.85\columnwidth]{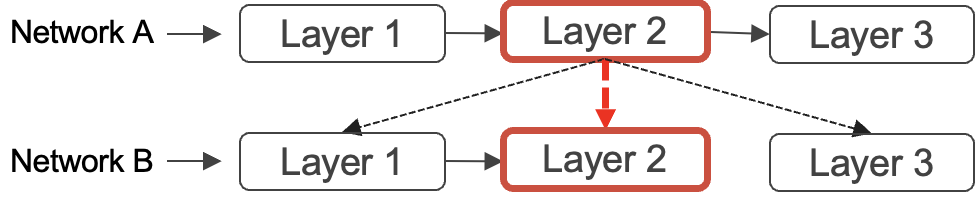}
\caption{Layer prediction benchmark. Given two models differing only in weight initialization, \mbox{A and B}, for each layer in the first model, among all layers of the second model, the highest similarity should be assigned to the architecturally-corresponding layer.}
\label{fig:layer_prediction}
\end{figure}

Proposed by \citet{kornblith2019similarity},  a basic and intuitive benchmark is to assess the invariance of a similarity measure against changes to a random seed. Given two models
differing only in their weight initializations, for each layer in the first model, among all layers of the second model, one can expect that a good similarity measure assigns the highest similarity for the architecturally-corresponding layer. Formally, let $f$ and $g$ be two models with $k$ layers, and define  $f_i$ and $g_i$ as the $i^{th}$ layer of $f$ and $g$ models, respectively. After calculating the similarity of $f_i$ to each layer of $g$ ($g_1, \ldots, g_k$), the pair with the highest similarity is expected to be $(f_i, g_i)$. This benchmark counts the number of layers for which this pair was indeed the most similar, and divides by the total number of pairs. An illustration is found in Figure \ref{fig:layer_prediction}.

The intuition behind this benchmark is that each layer captures different information about the input data. For example, \citet{jawahar2019does} showed that different layers of the BERT model capture different semantic information.
\subsection{Multilingual benchmark}

Multilingual models, such as Multilingual-BERT \citep{devlin2018bert}, learn to represent texts in different languages in the same representation space. Interestingly, these models show cross-lingual zero-shot transferability, where a model is fine-tuned in one language and evaluated in a different language \citep{pires2019multilingual}. \citet{muller2021first} analyzed this transferability and found that lower layers of the Multilingual-BERT align the representations between sentences in different languages.

\begin{figure}[t]
\centering
\includegraphics[width=0.95\columnwidth]{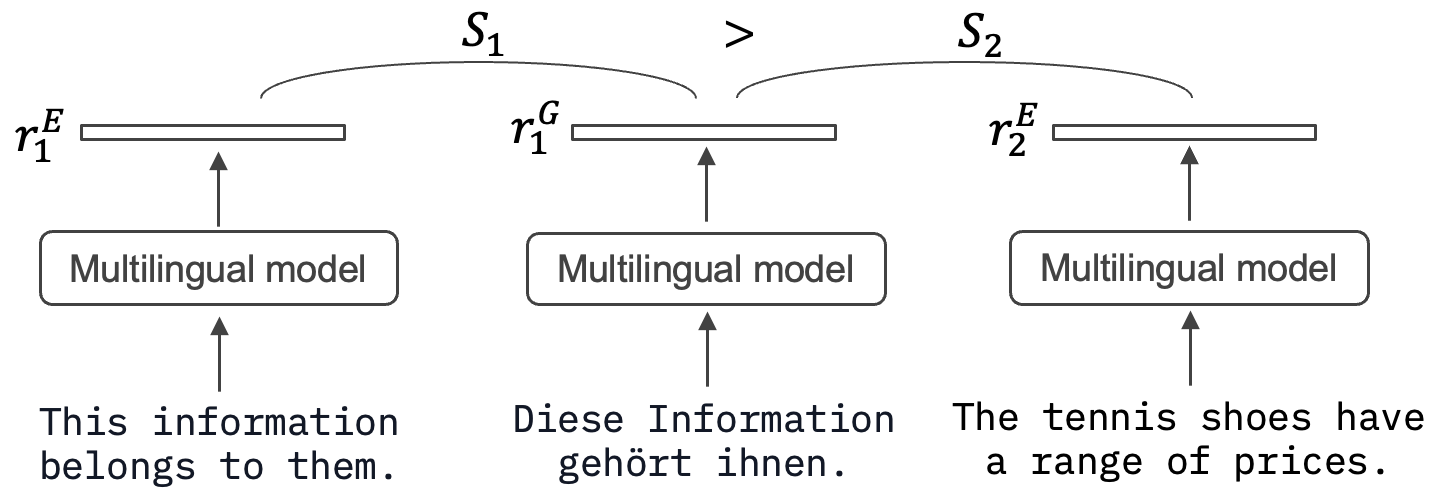}
\caption{The multilingual benchmark. $r_1^E$ and $r_1^G$ denote the representations of the same sentence in different languages, and $S_1$ is their similarity. $r_2^E$ represents the random sentence representation, and $S_2$ is the similarity between it and $r_1^G$. We expect $S_1$ to be higher than $S_2$.}
\label{fig:multilingual}
\end{figure}

Since multilingual models share similarities between representations of different languages, we expect that a good similarity measure should assign a high similarity to two representations of a sentence in two different languages. In other words, we expect similarity measures to be invariant to the sentence source language. Consider a multilingual model $f$ and dataset $\mathbb{X}$, where each entry consists of the same sentence in different languages.
Let $(x_1^{(i)}, x_2^{(i)}) \in \mathbb{X}$ be a sentence written in two languages -- language A and language B. The similarity between $f(x_1^{(i)})$ and $f(x_2^{(i)})$ should be higher than the similarity between $f(x_1^{(i)})$ and the representation of a sentence in language B randomly chosen from $\mathbb{X}$, i.e., $f(x_2^{(j)})$, where $(x_1^{(j)}, x_2^{(j)}) \in \mathbb{X}$ is a randomly chosen example from $\mathbb{X}$. 
The benchmark calculates the fraction of cases for which the correct translation was assigned the highest similarity. 
An illustration is found in Figure \ref{fig:multilingual}.

 Additionally, we suggest a strengthened version of the multilingual benchmark, using FAISS.
 Instead of sampling random sentences in language B, we use FAISS to find the pair $(x_1^{(j)}, x_2^{(j)}) \in \mathbb{X}$, where $x_2^{(j)} \neq x_2^{(i)}$, with the  representation $f(x_2^{(j)})$ that is most similar to $f(x_2^{(i)})$, out of a large set of vectors pre-indexed by FAISS.
 This leads to a more challenging scenario, as the similarity between $x_1^{(i)}$ and FAISS-sampled $x_2^{(j)}$ is expected to be higher than the similarity between $x_1^{(i)}$ and randomly chosen $x_2^{(j)}$, increasing the difficulty of identifying the pair $(x_1^{(i)}, x_2^{(i)})$ as the highest-similarity pair. Note that FAISS only affects the evaluation step, and is not used during \method{}'s training.

\subsection{Image-caption benchmark}
 Let $\mathbb{X}$ be a dataset of images and their textual descriptions (captions),  $f$ be a computer vision model and $g$ a language model.
Given a pair of an image and its caption, $(m^{(i)}, c^{(i)}) \in \mathbb{X}$, a good similarity measure is expected to assign a high similarity to their representations -- $f(m^{(i)}), g(c^{(i)})$. In particular, this similarity should be higher than that of the pair of the same image representation $f(m^{(i)})$ and some random caption's representation $g(c^{(j)})$, where $c^{(j)}$ is a randomly chosen caption from $\mathbb{X}$. 

\begin{figure}[t]
\centering
\includegraphics[width=0.85\columnwidth]{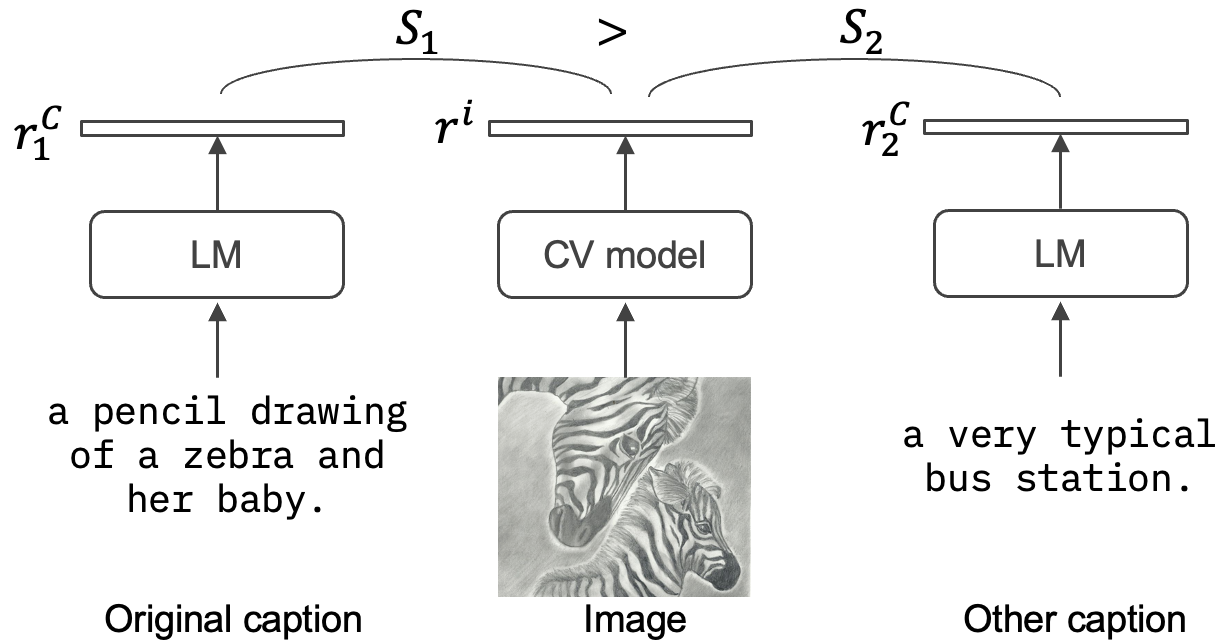}
\caption{The image--caption benchmark. $r_1^C$ and $r^i$ denote the representations of the caption and the image pair, respectively, and $S_1$ is their similarity. $r_2^C$ denotes the random caption representation, and $S_2$ is the similarity between it and $r^i$. $S_1$ should be greater than $S_2$.}
\label{fig:image_caption_illustration}
\end{figure}
The intuition behind this benchmark is that an image and its caption represent the same scene in a different way. Thus, their representations should have a higher similarity than that of the same image and some random caption. An illustration is found in Figure \ref{fig:image_caption_illustration}. As in the multilingual benchmark, we also propose a strengthened variant for the image--caption benchmark using  FAISS. Rather than sampling a random caption $c^{(j)}$, we use FAISS to find the pair $(m^{(j)}, c^{(j)}) \in \mathbb{X}$, where $c^{(j)} \neq c^{(i)}$, with the representation $g(c^{(j)})$ that is most similar to $g(c^{(i)})$. 


\section{Experiments}
\label{sec:expreriments}
\paragraph{Baselines.}

We compare \method{} with the following standard baselines.
\begin{itemize}
    \item \textbf{Centered Kernel Alignment (CKA)}: Proposed by \citet{kornblith2019similarity}, CKA computes a kernel matrix for each matrix representation input, and defines the scalar similarity index as the two kernel matrices' alignment. 
    We use a linear kernel for CKA evaluation, as the original paper reveals similar results for both linear and RBF kernels. CKA is our main point of comparison due to its success in prior work and wide applicability. 
    
    \item \textbf{PWCCA}: Proposed by \citet{morcos2018insights}, PWCCA is an extension of Canonical Correlation Analysis (CCA). Given two matrices, CCA finds bases for those matrices, such that after projecting them to those bases the correlation between the projected matrices  is maximized. While in CCA the scalar similarity index is computed as the mean correlation coefficient, in PWCCA that mean is weighted by the importance each canonical correlation has on the representation.\footnote{PWCCA and SVCCA require the number of examples to be larger than the feature vector dimension, which is not possible to achieve in all benchmarks. Therefore, we compare with them in a subset of our experiments.} 
\end{itemize}


See Appendix \ref{appendix:sim_measures} for more details on the current methods. In the main body we report results of the more successful methods: CKA and PWCCA. Additional baseline are reported in Appendix \ref{sec:additional_methods}.

\paragraph{Ablations.}
In addition, we report the results of  two new similarity measures, which use an encoder to map representations to the space where similarity is measured. However, 
in both methods we train $e_
{\theta}$ to only maximize the similarity between positive pairs:
\begin{equation}
\mathcal{L}_{max} =  -s(\bm{z}_1, \bm{z}_2)
\label{cl_max}
\end{equation}
where $\bm{z}_1$ and $\bm{z}_2$ are representations whose similarity we wish to maximize.  
We experiment with two functions for  $s$---dot-product and CKA---and accordingly name these similarity measures DeepDot  and  DeepCKA.  These methods provide a point of comparison where the similarity measure is trained, but \emph{without negative examples}, in order to assess the importance of contrastive learning. 
\paragraph{Encoders details.} In all experiments, the encoder $e_{\theta}$ is a two-layer multi-layered perceptron with hidden layer dimensions of $512$ and $256$, and output dimension of $128$.  We trained the encoder for 50 epochs for the layer prediction and 30 epochs for the multilingual and image--caption benchmarks. We used the Adam optimizer \citep{kingma2014adam} with a learning rate of $0.001$ and a batch size of $1024$ representations. We used $\tau = 0.07$ for \method{} training.

\subsection{Layer prediction benchmark}

\begin{table*}[t]
\centering
\begin{tabular}{l rr | l rr}
\toprule
 \multicolumn{3}{c}{Language} & \multicolumn{3}{c}{Vision} \\ 
\cmidrule(lr){1-3} \cmidrule(lr){4-6}
& Penn TreeBank & WikiText &  & CIFAR-10 & CIFAR-100 \\ \midrule
PWCCA                                                      & 38.33       & 55.00 & PWCCA  & 47.27 & 45.45 \\ 
CKA                                                        & 71.66       & 76.66 & CKA  & 78.18  & 74.54    \\
DeepDot                                                    & 15.55 \small $\pm$ 1.69       & 14.00 \small $\pm$ 2.26 & DeepDot  & 14.90 \small $\pm$ 1.78 & 14.18 \small $\pm$ 2.67  \\
DeepCKA                                                    & 16.66 \small $\pm$ 3.16       & 19.66 \small $\pm$ 1.63 & DeepCKA  & 17.09 \small $\pm$ 2.95 & 13.09 \small $\pm$ 4.20  \\
\midrule 
\method{} & & & \method{} & & \\ 
\hspace{.5em} Penn  & 100 \small $\pm$ 0         & 85.45 \small $\pm$ 1.62 & \hspace{.5em} CIFAR-10  & 100 \small $\pm$ 0   & 90.54 \small $\pm$ 2.90 \\ 
\hspace{.5em} Wiki  & 94.00 \small $\pm$ 4.66       & 100 \small $\pm$ 0 & \hspace{.5em} CIFAR-100  & 85.81 \small $\pm$ 5.68   & 100 \small $\pm$ 0  \\  
\bottomrule
\end{tabular}
\caption{Layer prediction benchmark accuracy results for language and vision cases. For encoder-based methods we report mean and std over 5 
 random initializations. For \method{}, we experiment with training with different datasets (rows) and evaluating on same or different datasets (columns).}
 \label{tab:layer_prediction}
\end{table*}

\subsubsection{Setup} 
Recall that this benchmark evaluates whether a certain layer from one model is deemed most similar to its architecturally-corresponding layer from another model, where the two models differ only in their weight initialization. We repeat this process for all layers and 5 different model pairs, and report average accuracy. 
We experiment with both language and vision setups.

\paragraph{Models.} For language experiments, we use the MultiBERTs \citep{sellam2021multiberts}, a set of 25 BERT models, differing only in their weights initialization. For vision experiments, we pre-train 10 visual transformer (ViT) models \citep{dosovitskiy2020image} on the ImageNet-1k dataset \citep{imagenet15russakovsky}. Then we fine-tune them on CIFAR-10 and CIFAR-100 datasets \citep{krizhevsky2009learning}.  Further details are available in Appendix \ref{appendix:vit_parameters}.

\paragraph{Datasets.} In language experiments, we use word-level contextual representations generated on two English text datasets: the Penn Treebank \citep{marcus-etal-1993-building} and WikiText \citep{merity2016pointer}. 
For Penn TreeBank we generate 5005/10019 test/training representations, respectively; for WikiText we generate  5024/10023 test/training representations. Vision experiments are conducted using representations generated on CIFAR-10 and CIFAR-100. For both we generate 5000 and 10000 test and training representations, respectively.

\paragraph{Positive and Negative sets.} Given a batch of representations of some model $i$ at layer $j$, we define its positive set as the representations at the same layer $j$ of all models that differ from $i$. The negative set is all representations from layers that differ from $j$ (including from model $i$).

\subsubsection{Results}
The results are shown in Table \ref{tab:layer_prediction}. In both language and vision evaluations, CKA achieves better results than PWCCA, consistent with the findings by  \citet{ding2021grounding}. 
 DeepDot and DeepCKA  perform poorly, with much lower results than PWCCA and CKA, revealing that maximizing the similarity is not satisfactory for similarity measure purposes. 
Our method, \method{}, achieves excellent results. When trained on one dataset's training set and evaluated on the same dataset's test set, \method{} achieves perfect accuracy under this benchmark, with a large margin over CKA results. This holds for both language and vision cases. Even when trained on one dataset and evaluated over another dataset, \method{}  surpasses other similarity measures, showing the transferability of the learned encoder projection between datasets. This is true both when transferring across domains (in text, between news texts from the Penn Treebank and Wikipedia texts), and when transferring across classification tasks (in images, between the 10-label CIFAR-10 and the 100-label CIFAR-100). 

\subsection{Multilingual benchmark}
\label{sec:multilingual}
\subsubsection{Setup}
This benchmark assesses whether a similarity measure assigns a high similarity to multilingual representations of the same sentence in different languages. Given a batch of (representations of) sentences $b^{(i)}$ in language $L_i$ and their translations $b^{(j)}$ in language $L_j$, we compute the similarity between $b^{(i)}$ and $b^{(j)}$, and the similarities between $b^{(i)}$ and 10 randomly chosen batches of representations in language $L_j$. If  $b^{(i)}$ is more similar to $b^{(j)}$ than  to all other batches, we mark success. (Alternatively, in a more challenging scenario, we use FAISS to find for each representation in each layer the 10 most similar representations in that layer.) We repeat this process separately for representations from different layers of a multilingual model, over many sentences and multiple language pairs, and report average accuracy per layer.\footnote{For deep similarity measures (DeepCKA, DeepDot, and \method{}), upon training the encoder on examples from a pair of languages, $(L_r, L_q), r \neq q$, we evaluate it over all other distinct pairs of languages.} Appendix \ref{appendix:multilingual} gives more details.

\begin{table*}[t]
\centering
\resizebox{\textwidth}{!}{\begin{tabular}{@{}ccccc|cccc@{}}
\toprule 
& \multicolumn{4}{c}{Random}                                                        & \multicolumn{4}{c}{FAISS}                                                   \\ \cmidrule(lr){2-5} \cmidrule(lr){6-9}
Layer & CKA              & DeepCKA          & DeepDot           & ContraSim        & CKA              & DeepCKA          & DeepDot           & ContraSim         \\ \midrule
1     & 71.7 \small $\pm$ 5.3 & 82.0 \small $\pm$ 6.4 & 63.3 \small$\pm$ 10.4 & 95.5 \small $\pm$ 5.4 & 20.1 \small $\pm$ 4.0 & 10.7 \small $\pm$ 2.6 & 29.9 \small  $\pm$ 8.7  & 36.0 \small $\pm$ 10.7 \\
2     & 78.7 \small $\pm$ 4.4 & 86.4 \small $\pm$ 4.1 & 68.5 \small $\pm$ 9.9 & 95.0 \small $\pm$ 7.2 & 27.2 \small $\pm$ 5.5 & 12.3 \small $\pm$ 2.9 & 46.9 \small  $\pm$ 9.8  & 33.0 \small $\pm$ 14.8 \\
3     & 86.8 \small $\pm$ 3.0 & 87.1 \small $\pm$ 3.2 & 70.4 \small $\pm$ 9.7 & 96.4 \small $\pm$ 6.7 & 41.9 \small $\pm$ 8.7 & 17.6 \small $\pm$ 4.2 & 51.5 \small $\pm$ 10.3  & 45.4 \small $\pm$ 20.5 \\
4     & 92.6 \small $\pm$ 1.4 & 91.5 \small $\pm$ 2.4 & 95.4 \small $\pm$ 3.4 & 99.9 \small $\pm$ 0.2 & 33.4 \small $\pm$ 7.0 & 15.2 \small $\pm$ 3.7 & 52.2 \small  $\pm$ 8.6  & 72.4  \small $\pm$ 9.8  \\
5     & 88.3 \small $\pm$ 3.2 & 83.5 \small $\pm$ 5.2 & 94.7 \small $\pm$ 4.8 & 99.9 \small $\pm$ 0   & 49.3 \small $\pm$ 4.3 & 36.9 \small $\pm$ 6.3 & 42.4 \small $\pm$ 12.9  & 99.1. \small $\pm$ 0.8  \\
6     & 88.6 \small $\pm$ 3.4 & 86.4 \small $\pm$ 5.2 & 92.5 \small $\pm$ 5.4 & 100  \small $\pm$ 0   & 51.4 \small $\pm$ 5.5 & 39.9 \small $\pm$ 7.2 & 42.1 \small $\pm$ 12.3  & 99.5. \small $\pm$ 0.4  \\
7     & 88.8 \small $\pm$ 3.7 & 86.9 \small $\pm$ 5.0 & 92.6 \small $\pm$ 5.0 & 100  \small $\pm$ 0   & 53.0 \small $\pm$ 5.8 & 41.1 \small $\pm$ 7.7 & 45.7 \small $\pm$ 11.7  & 99.6. \small $\pm$ 0.3  \\
8     & 89.3 \small $\pm$ 3.6 & 85.2 \small $\pm$ 5.7 & 91.4 \small $\pm$ 7.0 & 100  \small $\pm$ 0   & 56.1 \small $\pm$ 5.8 & 45.0 \small $\pm$ 8.7 & 43.8 \small $\pm$ 13.4  & 99.7. \small $\pm$ 0.3  \\
9     & 88.1 \small $\pm$ 3.8 & 82.4 \small $\pm$ 5.6 & 89.1 \small $\pm$ 9.5 & 100  \small $\pm$ 0   & 53.3 \small $\pm$ 4.9 & 42.7 \small $\pm$ 8.5 & 39.2 \small $\pm$ 12.9  & 99.6. \small $\pm$ 0.3  \\
10    & 87.0 \small $\pm$ 3.5 & 80.3 \small $\pm$ 5.9 & 85.3\small $\pm$ 10.3 & 100  \small $\pm$ 0   & 51.5 \small $\pm$ 5.3 & 42.4 \small $\pm$ 7.8 & 34.3 \small $\pm$ 12.2  & 99.5. \small $\pm$ 0.4  \\
11    & 86.7 \small $\pm$ 4.2 & 76.6 \small $\pm$ 6.4 & 79.7\small $\pm$ 13.9 & 99.9 \small $\pm$ 0 & 52.4 \small $\pm$ 5.3 & 43.3 \small $\pm$ 8.5 & 31.4 \small $\pm$ 12.8  & 99.3. \small $\pm$ 0.5  \\
12    & 86.4 \small $\pm$ 3.4 & 63.8 \small $\pm$ 7.9 & 64.3\small $\pm$ 19.7 & 99.9 \small $\pm$ 0 & 52.8 \small $\pm$ 4.5 & 32.3 \small $\pm$ 8.7 & 26.1 \small $\pm$ 21.9  & 98.9. \small $\pm$ 0.8  \\ 
\bottomrule
\end{tabular}}
\caption{Multilingual benchmark accuracy results. With random sampling (left block), \method{} outperforms other similarity measures. Using FAISS (right block) further extends the gaps.}
\label{tab:multilingual_results}
\end{table*}

\paragraph{Model and Data.} We use two multilingual models: multilingual BERT \citep{devlin2018bert}\footnote{https://huggingface.co/bert-base-multilingual-cased} and XLM-R \citep{conneau2019unsupervised}. 
We use the XNLI dataset \citep{conneau2018xnli}, which has natural language inference examples, parallel in multiple languages. Each example in our dataset is a  sentence taken from either the premise or hypothesis sets. We experiment with 5 typologically-different languages: English, Arabic, Chinese, Russian, and Turkish. We created sentence-level representations, with 5000 test 10000 training  representations.
As a sentence representation, we experiment with [CLS] token representations and with mean pooling of token representations, since \citet{del2021establishing} noted a difference in similarity in these two cases. We report results with [CLS] representations in the main body and with mean pooling in Appendix \ref{appendix:multilingual}; the trends are similar. 

\paragraph{Positive and Negative sets.} 
Given a pair of languages and a batch of representations at some layer, for each representation we define its positive pair as the representation of the sentence in the different language, and its negative set as all other representations in the batch.

\subsubsection{Results}
Results with multilingual BERT representations in Table \ref{tab:multilingual_results} show our method's effectiveness. (Trends with XLM-R are consistent; Appendix~\ref{app:xlmr}). Under random sampling evaluation (left block), \method{} shows superior results over other similarity measures, despite being evaluated on language pairs it hasn't seen at training. Using FAISS sampling (right block) further extends the gaps. While CKA results dropped by $\approx 45\%$, DeepCKA dropped by $\approx51\%$, and DeepDot dropped by $\approx 40\%$, \method{} was much less affected by FAISS sampling ($\approx 17\%$ drop on average and practically no drop in most layers). This demonstrates the high separability between examples of \method{}, enabling it to distinguish even very similar examples.
For all other methods, mid-layers have the highest accuracy, whereas for our method almost all layers are near $100\%$ accuracy, except for the first 3 or 4 layers. 

\begin{figure}[h]
\centering
\includegraphics[width=0.46\textwidth]{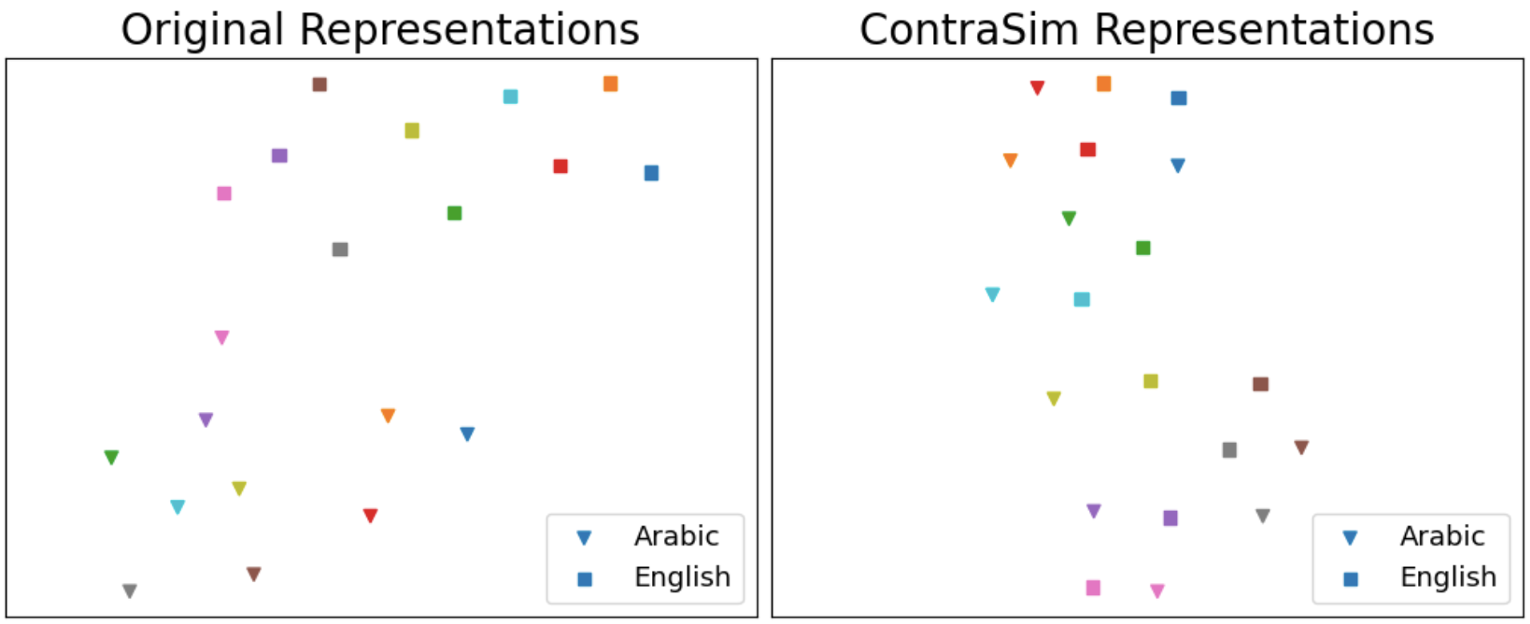}
\caption{ 
Original representations (left) are clustered by the source language (by shape). 
\method{}  (right) projects representations of the same sentence in different languages close by  (by color).}
\label{fig:multilingual_embeddings}
\end{figure}

To further analyze this, we compare the original multilingual representations from the last layer with their projections by \method{}'s trained encoder. 
Figure \ref{fig:multilingual_embeddings} shows UMAP \citep{mcinnes2018umap} projections for 10 English sentences and 10 Arabic sentences, before and after \method{} encoding.  The \method{} encoder was trained on Arabic and English languages. The original representations are organized according to the source language (by shape), whereas \method{} projects translations of the same sentence close to each other (clustered by color).

\subsection{Image--caption benchmark}

\begin{figure*}[t]
\centering
\includegraphics[width=0.64\linewidth]{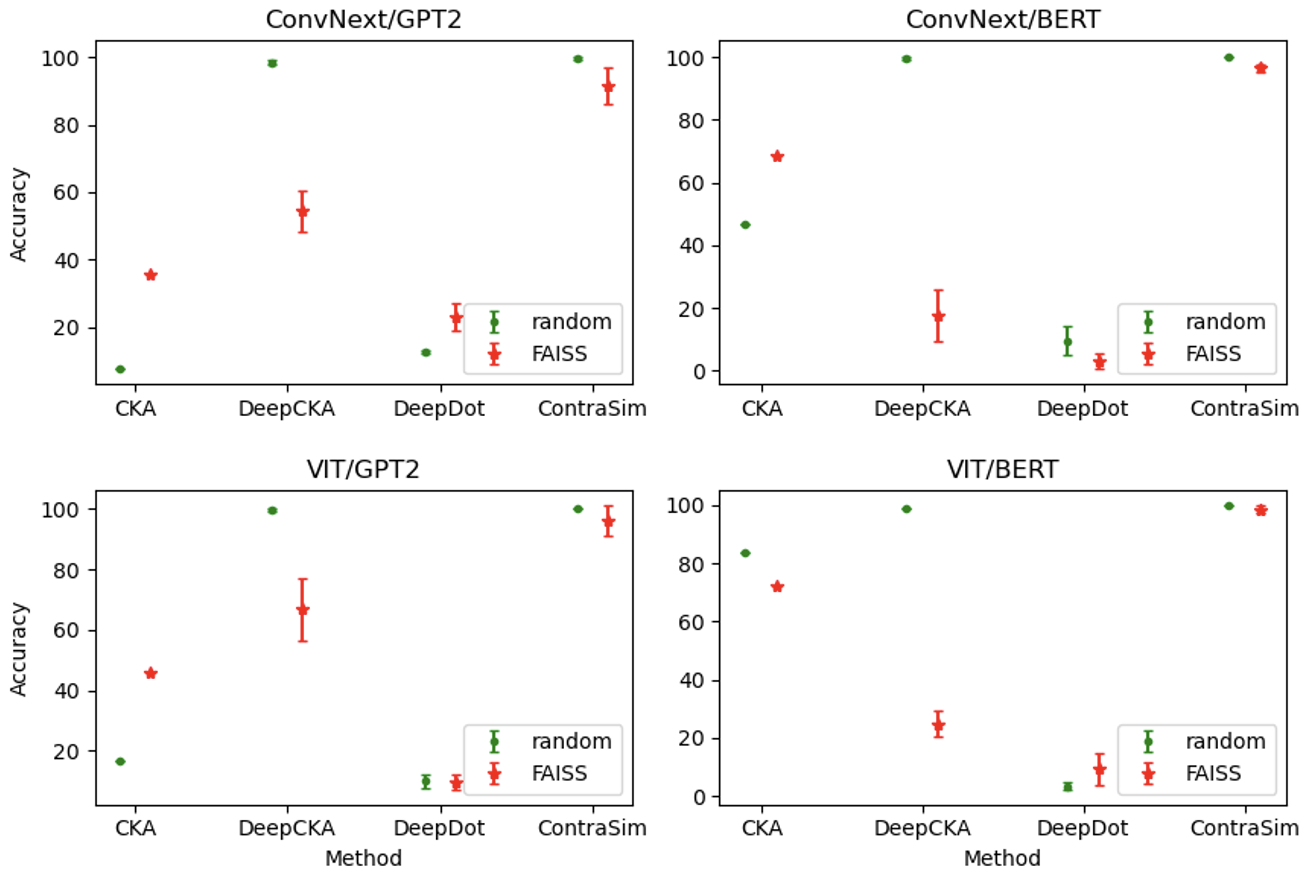}
\caption{Image--caption benchmark results for 4 different model pairs. \method{} works best, and is the only measure robust to FAISS sampling.}
\label{fig:image_caption}
\end{figure*}

\subsubsection{Setup}

Given a test set $\mathbb{X}$, consisting of pairs of an image representation generated by a CV model and its caption representation from a LM, we split  $\mathbb{X}$ to batches of size 64. For each batch, we compute the similarity between the image representations and their corresponding caption representations. We then sample 10 different caption batches, either randomly or using FAISS (as before), and compute the similarity between the image representation and each random/FAISS-retrieved caption representation. If the highest similarity is between the image representation and the original caption representation, we mark a success. 
For trainable similarity measures, we train with 5 different random seeds and average the results.

\paragraph{Models and Data.} 
We use two vision models for image representations: ViT  and ConvNext \citep{liu2022convnet}; and two  language models for text representations: BERT  and GPT2 \citep{radford2019language}.
%
We use the Conceptual Captions dataset \citep{sharma2018conceptual}.
We use 5000 and 10000 pairs as test and training sets, respectively.

\paragraph{Positive and Negative sets.}
Given a batch of image representations with their corresponding caption representations, for each image representation we define as a positive set its corresponding caption representation, and as a negative set all other representations in the batch.

\subsubsection{Results}
Figure \ref{fig:image_caption} demonstrates the strength of \method{}. Under random sampling (green boxes),  DeepCKA achieves comparable results to \method{}, while DeepDot and CKA achieve lower results. 
However, using FAISS (red boxes) causes a big decrease in DeepCKA accuracy, while \method{} maintains high accuracy. 
Furthermore, in 3 of 4 pairs we tested, FAISS sampling yielded better CKA accuracy than random sampling. This contradicts the intuition that similar examples at the sampling stage should make it harder for similarity measures to distinguish between examples. This might indicate that CKA suffers from stability issues.
Finally, we report results with the multi-modal CLIP model \citep{radford2021learning} in Table \ref{tab:image_caption_clip} (Appendix \ref{app:image_caption}). Because the model was pre-trained with contrastive learning, simple dot-product similarity works very well, so there is no need to learn a similarity measure in this case.  

\subsubsection{\method{} on different dimensions}
\label{different_dim}
We further evaluated \method{} with different representation dimensions. We performed the same image--caption benchmark with the exception that we used different vision model sizes: ViT-large and ConvNext-base, both with 1024-dimensional representation vectors. As language models we used the same GPT2 and BERT models, with a 768-dimensional representation vector. 

\begin{table}[]
\centering
\begin{tabular}{@{}ccc@{}}
\toprule
                   & CKA   & ContraSim \\ \midrule
ViT-large/GPT2     & 45.57 & 98.73     \\
ViT-large/BERT     & 83.54 & 98.73     \\
ConvNext-base/GPT2 & 39.24 & 100       \\
ConvNext-base/BERT & 74.68 & 98.73     \\ \bottomrule
\end{tabular}
\caption{Image--caption benchmark accuracy results for model pairs with different dimensions. We report results using FAISS sampling. Despite different model dimensions, \method{} consistently works best.}
\label{tab:different_dim}
\end{table}

We trained a different encoder for each model, as opposed to the single encoder we trained in all other experiments. This enables \method{} to be used with representations with different dimensions. Results are summarized in Table \ref{tab:different_dim}. We report results with FAISS sampling. Across all pairs, \method{} achieves superior results.

\section{Interpretability insights}

Having shown the superiority of our method, we now discuss a few interpretability insights that arise from our evaluations, and are not revealed by previous similarity measures. 

In the multilingual benchmark (Table~\ref{tab:multilingual_results},  FAISS results),  we found a much greater difference in accuracy between shallow and deep layers in \method{} compared to previous similarity measures. Using previous similarity measures we might infer that there is no difference in the ability to detect the correct pairs across different layers. However, \method{} shows that the difference in the ability to detect the correct pair dramatically changes from shallow to deep layers. This raises an interesting insight regarding the evolution of representations across layers. For instance, \citet{conneau2020emerging} used CKA to measure the similarity of representations of bilingual models of different languages and the same language (using back-translation). From their results it can be observed that there is no much difference in similarity of sentences from different languages and from the same language at the shallow and deep layers. Our results show that this difference is higher than found before.

In the image--caption benchmark (Figure~\ref{fig:image_caption}), from the CKA results we might infer that
BERT representations are more similar to computer vision representations than GPT2 representations. That is because with CKA, it is easier to detect the matching image--caption pair with BERT than it is with GPT2.  However, \method{} achieves a high accuracy in both BERT pairs and GPT2 pairs, which means that both language models about as similar to  vision models, in contrast to what we may infer from previous similarity measures. This reveals a new understanding regarding the relationship between language  and vision models. To the best of our knowledge, no prior work has done such a similarity analysis.

\section{Conclusion}
We proposed a new similarity measure for interpreting neural networks, \method{}. By defining positive and negative sets we learn an encoder that maps representation to a space where similarity is measured. 
Our method outperformed other similarity measures under the common layer prediction benchmark and two new benchmarks we proposed: the multilingual benchmark and the image--caption benchmark. It particularly shines in strengthened versions of said benchmarks, where random sampling is replaced with finding the most similar examples using FAISS. Moreover, we show that even when \method{} is trained on data from one domain/task and evaluated on data from another domain/task, it achieves superior performance. 
 Considering \method{}'s superiority in all evaluations, we believe it is a better tool for the interpretability of neural networks, and have discussed a few insights revealed by \method{} and not captured by previous methods.

Our new similarity measure benchmarks can facilitate work on similarity-based analyses of deep networks. The multilingual benchmark is useful for work on multilingual language models, while the image--caption benchmark may help in multi-modal settings. 
In addition, since our method learns a parameterized measure, it may help train models with similarity objectives.

\section{Limitations}
Compared to existing methods, \method{} needs access to a training set for the encoder training procedure. The training procedure itself is efficient, typically a matter of minutes.

\section{Ethics Statement} 
Our work adds to the body of literature on the interpretability of neural networks and may mitigate their opacity. We do not foresee major risks associated with this work. However, a malicious actor could train \method{} adversarially, assign poor similarity estimates, and lead to false analyses.

\section*{Acknowledgements}
This work was supported by an
AI alignment grant from Open Philanthropy, the Israel Science Foundation (grant No. 448/20), and an Azrieli Foundation Early Career Faculty Fellowship.
\bibliography{custom}

\clearpage
\appendix
\section{Appendix}
\label{sec:appendix}

\subsection{Multilingual benchmark}
\label{appendix:multilingual}
\subsubsection{Evaluation parameters}
We split the test set, $\mathbb{X}$, into equally sized batches of size 8, $\{b^{(1)}, b^{(2)}, ..., b^{(n )}\}$, where each batch consists of multilingual BERT representations  of the same sentence in 5 different languages: $L=\{L_1, ..., L_5\} $.  Given a pair of different languages, $(L_i, L_j), i \neq j$, and a batch of representations, $b$, we consider the representation of those languages in the batch, $(b[i], b[j])$, and compute the similarity between $b[i]$ and $b[j]$ as $s_0 \equiv s(b[i], b[j])$. We also compute the similarity between $b[i]$ and 10 randomly chosen batches (or, 10 batches chosen using FAISS) of representations in language $L_j$ as $\{s_t \equiv s(b[i], b_t[j])\}_{t=1}^{10}$. 
If $\argmax_t \{s_t\}_{t=0}^{10} = 0$, we count it as a correct prediction. Each layer's accuracy is defined as the number of successful predictions over the number of batches, $n$. We average results over all possible pairs of different languages.  

\subsubsection{Further evaluations} \label{app:xlmr}
In addition to using the [CLS] token representation as a sentence representation, we also evaluated the multilingual benchmark using mean pooling sentence representation. We used the same evaluation process as described in Section \ref{sec:multilingual}. The results, summarized in Table \ref{tab:multilingual_mean}, are consistent with the results in the main paper (Table \ref{tab:multilingual_results}). Under random sampling, \method{} outperforms all other similarity measures. Using FAISS causes a big degradation in all other methods' accuracy, while \method{} maintains a high accuracy across all layers. 

\label{sec:multilingual_mean}

\begin{table*}[t]
\centering

\resizebox{\textwidth}{!}{\begin{tabular}{@{}ccccc|cccc@{}}

      & \multicolumn{4}{c}{Random}                                                                                                                          & \multicolumn{4}{c}{FAISS}                                                   \\ \midrule
Layer & CKA                           & DeepCKA                       & DeepDot                        & \multicolumn{1}{c|}{ContraSim}                     & CKA                            & DeepCKA                        & DeepDot                        & ContraSim                      \\ \midrule
1     & 87.7 \small $\pm$ 6.9 & 86.3  \small $\pm$ 9.7 & 43.4 \small $\pm$ 17.7 & 98.7 \small $\pm$ 2.2 & 67.6 \small $\pm$ 14.3 & 54.1 \small $\pm$ 10.9  & 41.7 \small $\pm$ 19.1 & 94.2 \small $\pm$ 10.7 \\
2     & 89.0 \small $\pm$ 6.3 & 88.7  \small $\pm$ 7.0 & 51.5 \small $\pm$ 20.0 & 99.5 \small $\pm$ 0.8 & 68.2 \small $\pm$ 13.2 & 49.0  \small $\pm$ 8.3  & 38.9 \small $\pm$ 17.2 & 96.6 \small $\pm$ 14.8 \\
3     & 91.8 \small $\pm$ 4.4 & 90.7  \small $\pm$ 6.0 & 63.3 \small $\pm$ 20.8 & 99.9 \small $\pm$ 0.1 & 72.2 \small $\pm$ 11.5 & 55.4  \small $\pm$ 8.1  & 44.8 \small $\pm$ 16.6 & 98.8 \small $\pm$ 20.5 \\
4     & 93.7 \small $\pm$ 3.3 & 91.3  \small $\pm$ 5.0 & 73.1 \small $\pm$ 19.4 & 99.9 \small $\pm$ 0.0 & 74.3 \small $\pm$ 10.0 & 55.1  \small $\pm$ 8.0  & 45.7 \small $\pm$ 16.5 & 99.5 \small  $\pm$ 7.1  \\
5     & 95.3 \small $\pm$ 3.0 & 92.1  \small $\pm$ 4.0 & 83.9 \small $\pm$ 15.6 & 99.9 \small $\pm$ 0.0 & 78.2 \small  $\pm$ 8.2 & 56.7  \small $\pm$ 8.1  & 53.2 \small $\pm$ 17.5 & 99.8 \small  $\pm$ 4.4  \\
6     & 95.9 \small $\pm$ 2.4 & 91.8  \small $\pm$ 3.9 & 91.2 \small $\pm$ 10.6 & 100  \small $\pm$ 0   & 77.6 \small  $\pm$ 7.9 & 54.2  \small $\pm$ 8.1  & 60.1 \small $\pm$ 18.2 & 99.8 \small  $\pm$ 1.7  \\
7     & 95.4 \small $\pm$ 2.5 & 90.6  \small $\pm$ 4.1 & 93.1 \small  $\pm$ 9.2  & 100  \small $\pm$ 0  & 77.9 \small  $\pm$ 7.8 & 53.3  \small $\pm$ 7.2  & 63.5 \small $\pm$ 18.5 & 99.9 \small  $\pm$ 0.7  \\
8     & 94.8 \small $\pm$ 3.2 & 89.7  \small $\pm$ 4.3 & 90.3 \small $\pm$ 12.0 & 100  \small $\pm$ 0   & 76.7 \small  $\pm$ 8.1 & 52.4  \small $\pm$ 7.4  & 61.0 \small $\pm$ 19.8 & 99.9 \small  $\pm$ 0.3  \\
9     & 94.0 \small $\pm$ 3.4 & 88.5  \small $\pm$ 5.0 & 86.4 \small $\pm$ 15.1 & 100  \small $\pm$ 0   & 73.9 \small  $\pm$ 8.8 & 51.4  \small $\pm$ 7.8  & 55.5 \small $\pm$ 20.0 & 99.9 \small  $\pm$ 0.1  \\
10    & 92.6 \small $\pm$ 4.2 & 85.6  \small $\pm$ 5.9 & 80.7 \small $\pm$ 18.8 & 100  \small $\pm$ 0   & 72.2 \small  $\pm$ 8.4 & 49.3  \small $\pm$ 8.4  & 49.2 \small $\pm$ 20.6 & 99.9 \small  $\pm$ 0.1  \\
11    & 91.1 \small $\pm$ 5.1 & 81.0  \small $\pm$ 6.5 & 72.2 \small $\pm$ 23.7 & 99.9 \small $\pm$ 0   & 70.6 \small $\pm$ 10.1 & 48.8  \small $\pm$ 9.1  & 43.2 \small $\pm$ 20.7 & 99.8 \small  $\pm$ 0.1  \\
12    & 90.8 \small $\pm$ 5.8 & 71.3  \small $\pm$ 7.6 & 71.0 \small $\pm$ 21.0 & 99.9 \small $\pm$ 0   & 72.7 \small $\pm$ 11.3 & 40.3  \small $\pm$ 8.7  & 42.7 \small $\pm$ 17.0 & 99.4 \small  $\pm$ 0.1  \\ \bottomrule
\end{tabular}}
\caption{Multilingual benchmark results with mean pooling.}
\label{tab:multilingual_mean}
\end{table*}

In addition, we evaluated the multilingual benchmark with another multilingual model -- the XLM-R \citep{conneau2019unsupervised} model. Results, summarized in Table \ref{tab:multilingual_xlm}, show a similar pattern to Tables \ref{tab:multilingual_results} and \ref{tab:multilingual_mean}, with \method{} achieving the highest accuracies across all layers, in both random sampling and FAISS sampling scenarios. 

\begin{table*}[h]
\centering
\resizebox{\textwidth}{!}{\begin{tabular}{@{}ccccccccc@{}}
\multicolumn{5}{c}{Random} &
  \multicolumn{4}{c}{FAISS} \\ \midrule
Layer &
  CKA &
  DeepCKA &
  DeepDot &
  \multicolumn{1}{c|}{ContraSim} &
  CKA &
  DeepCKA &
  DeepDot &
  ContraSim \\ \midrule
1 &
  89.7 \small $\pm$. 4.6 &
  88.2 \small $\pm$  3.1 &
  45.0 \small $\pm$ 22.1 &
  \multicolumn{1}{c|}{99.89 \small $\pm$ 0.31} &
  60.6 \small $\pm$ 11.4 &
  24.6 \small $\pm$  4.7 &
  24.8 \small $\pm$ 12.4 &
  98.1 \small $\pm$  2.5 \\
2 &
  90.6 \small $\pm$  4.1 &
  92.3 \small $\pm$  2.0 &
  63.1 \small $\pm$ 23.4 &
  \multicolumn{1}{c|}{99.99 \small $\pm$ 0.02} &
  57.1 \small $\pm$ 11.4 &
  30.7 \small $\pm$  5.7 &
  31.1 \small $\pm$ 13.5 &
  99.5 \small $\pm$  0.7 \\
3 &
  92.8 \small $\pm$ 3.1 &
  93.8 \small $\pm$ 1.7 &
  79.5 \small $\pm$18.9 &
  \multicolumn{1}{c|}{99.99 \small $\pm$ 0} &
  55.5 \small $\pm$10.1 &
  33.8 \small $\pm$ 6.4 &
  39.3 \small $\pm$15.5 &
  99.9 \small $\pm$ 0.1 \\
4 &
  94.7 \small $\pm$ 2.6 &
  94.3 \small $\pm$ 1.7 &
  91.4 \small $\pm$11.7 &
  \multicolumn{1}{c|}{100 \small $\pm$ 0} &
  62.3 \small $\pm$ 9.5 &
  36.3 \small $\pm$ 6.8 &
  55.1 \small $\pm$16.2 &
  99.9 \small $\pm$ 0 \\
5 &
  95.9 \small $\pm$ 2.1 &
  94.6 \small $\pm$ 1.6 &
  94.0 \small $\pm$10.0 &
  \multicolumn{1}{c|}{100 \small $\pm$ 0} &
  66.2 \small $\pm$ 8.4 &
  37.2 \small $\pm$ 7.6 &
  64.2 \small $\pm$16.2 &
  99.9 \small $\pm$ 0 \\
6 &
  95.9 \small $\pm$ 2.1 &
  94.9 \small $\pm$ 1.6 &
  94.6 \small $\pm$ 8.9 &
  \multicolumn{1}{c|}{100 \small $\pm$ 0} &
  66.7 \small $\pm$ 8.3 &
  41.2 \small $\pm$ 7.6 &
  66.0 \small $\pm$17.5 &
  99.9 \small $\pm$ 0 \\
7 &
  96.6 \small $\pm$ 2.0 &
  94.9 \small $\pm$ 1.6 &
  94.9 \small $\pm$ 8.5 &
  \multicolumn{1}{c|}{100 \small $\pm$ 0} &
  71.7 \small $\pm$ 8.5 &
  44.1 \small $\pm$ 8.2 &
  68.8 \small $\pm$17.5 &
  99.9 \small $\pm$ 0 \\
8 &
  96.1 \small $\pm$ 2.1 &
  94.8 \small $\pm$ 1.7 &
  94.0 \small $\pm$ 9.3 &
  \multicolumn{1}{c|}{100 \small $\pm$ 0} &
  68.1 \small $\pm$ 8.3 &
  43.6 \small $\pm$ 7.9 &
  65.3 \small $\pm$18.0 &
  99.9 \small $\pm$ 0 \\
9 &
  94.8 \small $\pm$ 2.2 &
  94.9 \small $\pm$ 1.6 &
  93.3 \small $\pm$ 9.1 &
  \multicolumn{1}{c|}{100 \small $\pm$ 0} &
  58.5 \small $\pm$ 8.7 &
  42.8 \small $\pm$ 8.0 &
  61.7 \small $\pm$18.6 &
  99.9 \small $\pm$ 0 \\
10 & 
  93.9 \small $\pm$ 2.3 &
  94.3 \small $\pm$ 1.7 &
  92.7 \small $\pm$10.7 &
  \multicolumn{1}{c|}{100 \small $\pm$ 0} &
  46.3 \small $\pm$ 8.2 &
  39.1 \small $\pm$ 7.5 &
  56.6 \small $\pm$ 18.9 &
  99.9 \small $\pm$ 0 \\
11 &
  92.0 \small $\pm$  2.8 &
  93.3 \small $\pm$  2.2 &
  92.7 \small $\pm$ 10.9 &
  \multicolumn{1}{c|}{100 \small $\pm$ 0} &
  35.5 \small $\pm$  7.0 &
  39.1 \small $\pm$  7.3 &
  57.2 \small $\pm$ 18.3 &
  99.9 \small $\pm$  0 \\
12 &
  80.7 \small $\pm$ 4.7 &
  89.5 \small $\pm$ 3.0 &
  81.6 \small $\pm$ 13.8 &
  \multicolumn{1}{c|}{100 \small $\pm$ 0} &
  26.5  \small $\pm$  5.8 &
  32.3  \small $\pm$  7.0 &
  34.7  \small $\pm$ 15.1 &
  99.9  \small $\pm$  0 \\ \bottomrule
\end{tabular}}
\caption{Multilingual benchmark results on XLM-R model.}
\label{tab:multilingual_xlm}
\end{table*}

\subsection{Image--caption}
\label{app:image_caption}
In addition to the four model pairs we evaluated in Figure \ref{fig:image_caption}, we assessed the multi-modal vision and language CLIP model \citep{radford2021learning}, which was trained using contrastive learning on pairs of images and their captions. Results in Table \ref{tab:image_caption_clip} show interesting findings. Under random sampling, dot product, DeepCKA and \method{} achieve perfect accuracy. However, using FAISS causes significant degradation in DeepCKA accuracy, and only a small degradation in dot product and \method{} results, with equal accuracy for both. We attribute this high accuracy for simple dot product to the fact that CLIP training was done using contrastive learning, thus observing high separability between examples.
  
\begin{table}[h]
\centering
\begin{tabular}{l r r }
\toprule
                                                      & Random & FAISS \\ \midrule
CKA                                                   & 93.67  & 25.31 \\
Dot Product & 100    & 98.73 \\
DeepCKA                                               & 100    & 13.92 \\
DeepDot                                               & 29.11  & 25.31 \\
ContraSim                                             & 100    & 98.73 \\ \bottomrule
\end{tabular}
\caption{Image--caption benchmark accuracy results using CLIP model}
\label{tab:image_caption_clip}
\end{table}

\subsection{ViT training details}
\label{appendix:vit_parameters}
We used the ViT-base \citep{dosovitskiy2020image} architecture. We pretrained 10 models on the ImgaeNet-1K dataset \citep{deng2009imagenet}, differing only in their weight initializations by using random seeds from 0 to 9. We used the AdamW optimizer \citep{kingma2014adam} with $\text{lr}=0.001$, $\text{weight decay} = 1e-3$, $\text{batch size} = 128$, and a cosine learning scheduler. We trained each model for 150 epochs and used the final checkpoint.

Then, we fine-tuned the pretrained models on CIFAR-10 and CIFAR-100 datasets \citep{krizhevsky2009learning}. We used AdamW optimizer with $\text{lr}=2e-5$, $\text{weight decay} = 0.01$, $\text{batch size} = 10$, and a linear scheduler. For models fine-tuned on CIFAR-10, the average accuracy on the CIFAR-10 test set is $96.33\%$. For models fine-tuned on CIFAR-100, the average accuracy on the CIFAR-100 test set is $78.87\%$.

\subsection{Details of Prior Similarity Measures}
\label{appendix:sim_measures}
\paragraph{Canonical Correlation Analysis (CCA).}
 Given two matrices, CCA finds bases for those matrices, such that after projecting them to those bases the projected matrices' correlation is maximized. For $1 \leq i \leq p_1$, the $i^{\text{th}}$ canonical correlation coefficient $\rho_i$ is given by:
\begin{equation}
\begin{aligned}
\rho_i = \max_{w_X^i, w_Y^i}&  \text{corr}(Xw_X^i, Yw_Y^i)\\
\mathrm{s.t.} &\ ~~ \forall_{j< i}~~ Xw_X^i\perp Xw_X^j\\
&\ ~~ \forall_{j< i}~~ Yw_Y^i\perp Yw_Y^j .
\end{aligned}
\end{equation}
where $\mathrm{corr}(X, Y) = \frac{\langle X, Y \rangle}{\|X\| \cdot \|Y\|}$. Given the vector of correlation coefficients $\mathrm{corrs} = (\rho_1, ..., \rho_{p_1})$, the final scalar similarity index is computed as the mean correlation coefficient:
\begin{equation}
\begin{aligned}
S_{\mathrm{CCA}}(X, Y) = \overline{\rho}_{CCA} = \frac{\sum_{i=1}^{p_1} \rho_{i}}{p_1}
\end{aligned}
\end{equation}
as previously used in \citep{raghu2017svcca, kornblith2019similarity}.

\paragraph{Projection-Weighted CCA (PWCCA).} 
\citet{morcos2018insights} proposed a different approach to transform the vector of correlation coefficients, $\mathrm{corrs}$, into a scalar similarity index. Instead of defining the similarity as the mean correlation coefficient, PWCCA uses a weighted mean and the similarity is defined as:
\begin{align}
S_{PW} &= \frac{\sum_{i=1}^{p_1} \alpha_i \rho_i}{\sum_{i} \alpha_i} & 
\alpha_i &=\sum_{j} |\langle h_i, x_j \rangle|
\end{align}
where $x_j$ is the $j^\text{th}$ column of $X$, and $h_i = Xw_X^i$ is the vector observed upon projecting $X$ to the $i^{\text{th}}$ canonical direction. 

\paragraph{Singular Vector CCA (SVCCA).} Another extension to CCA, proposed by \citet{raghu2017svcca}, performs CCA on the truncated singular value decomposition (SVD) of the activation matrices. SVCCA keeps enough principal components to explain a fixed percentage of the variance.

Code available at: \url{https://github.com/google/svcca}.

\paragraph{Centered Kernel Alignment (CKA).} CKA, Proposed by \citet{kornblith2019similarity}, suggests computing a kernel matrix for each matrix representation input, and defining the scalar similarity index as the two kernel matrices' alignment. For linear kernel, CKA is defined as:
\begin{align}
S_{\mathrm{CKA}} = \frac{\|Y^TX\|_F^2}{\|X^TX\|_F\|Y^TY\|_F}
\end{align}

Code available at: \url{https://github.com/google-research/google-research/tree/master/representation_similarity}.

\paragraph{Norm.} For two representations, $x$ and $y$, we defined the dis-similarity measure as the norm of the difference between the normalized representations:

\begin{align}
    Dis_{\mathrm{Norm}}(x, y) = \|(x / \|x\| - y / \|y\|)\|
\end{align}

Since this is a dis-similarity measure, we defined the norm similarity measure as:

\begin{align}
    S_{\mathrm{Norm}} = 1 - Dis_{\mathrm{Norm}}(x, y)
\end{align}

For a batch of representations, we define batch similarity as the mean of pairwise norm similarity. 

\paragraph{Dot-product.} Measuring the similarity score between two feature vectors as their dot-product.

\subsection{Additional Evaluations}
\label{sec:additional_methods}

In this section, we report evaluation results with additional baselines. Furthermore, we evaluated \method{} with a different similarity measure than the dot-product and replaced it with the norm similarity measure.

\begin{table}[]
\centering
\begin{tabular}{@{}ccc@{}}
\toprule
                & Penn TreeBank & WikiText \\ \midrule
SVCCA           & 46.66         & 56.66    \\
Dot product     & 8.33          & 6.66     \\
Norm            & 10.00            & 11.66    \\ \midrule
ContraSim\_norm &               &          \\
Penn            & 100.00           & 90.00       \\
Wiki            & 100.00           & 100.00      \\ \bottomrule
\end{tabular}
\caption{Layer prediction benchmark with additional similarity measures.}
\label{tab:additional_layer_prediction}
\end{table}

Similar to PWCCA, SVCCA requires that the number of examples is larger than the vector dimension, thus we could only evaluate it in the layer prediction benchmark. All other similarity measures were evaluated with all evaluations - the layer prediction benchmark, the multilingual benchmark and the image-caption benchmark.

Table \ref{tab:additional_layer_prediction} shows layer prediction benchmark results. We can observe that SVCCA achieves slightly better results than PWCCA, and lower than CKA and \method{}. Both dot product and norm achieve low accuracies. ContraSim\_norm achieves the same or better results than \method{}.  

\begin{table*}[]
\centering
\begin{tabular}{@{}ccccccc@{}}
\toprule
&
  \multicolumn{3}{c}{Random} &
  \multicolumn{3}{c}{FAISS} 
   \\ \cmidrule(l){2-7} 
Layer &
  \multicolumn{1}{c}{\begin{tabular}[c]{@{}l@{}}Dot\\ product\end{tabular}} &
  Norm &
  \multicolumn{1}{c|}{ContraSim\_norm} &
  \begin{tabular}[c]{@{}c@{}}Dot\\ Product\end{tabular} &
  \multicolumn{1}{l}{Norm} &
  ContraSim\_norm \\ \midrule
1 &
  48.93 \small $\pm$ 7.07 &
  68.67 \small $\pm$ 10.86 &
  \multicolumn{1}{c|}{89.39 \small $\pm$ 14.46} &
  15.00 \small $\pm$ 6.41 &
  20.42 \small $\pm$ 9.32 &
  15.76 \small  $\pm$ 6.56 \\
2 &
  31.71 \small $\pm$ 3.31 &
  74.19 \small $\pm$ 9.73 &
  \multicolumn{1}{c|}{85.40 \small $\pm$ 18.00} &
  20.14 \small $\pm$ 2.75 &
  21.16 \small $\pm$ 11.43 &
  18.83 \small  $\pm$ 11.19 \\
3 &
  49.32 \small $\pm$ 6.53 &
  83.92 \small $\pm$ 13.12 &
  \multicolumn{1}{c|}{85.68 \small $\pm$ 18.88} &
  13.00 \small $\pm$ 4.50 &
  29.42 \small $\pm$ 22.99 &
  26.36 \small $\pm$ 16.01 \\
4 &
  29.60 \small $\pm$ 2.44 &
  99.61 \small $\pm$ 0.41 &
  \multicolumn{1}{c|}{96.81 \small $\pm$ 5.04} &
  16.20 \small $\pm$ 4.01 &
  57.98 \small $\pm$ 15.68 &
  28.79 \small  $\pm$ 9.80 \\
5 &
  99.75 \small $\pm$ 0.29 &
  99.86 \small $\pm$ 0.33 &
  \multicolumn{1}{c|}{99.86 \small $\pm$ 0.23} &
  82.17 \small $\pm$ 7.36 &
  82.39 \small $\pm$ 7.73 &
  74.04 \small $\pm$ 7.11 \\
6 &
  99.75 \small $\pm$ 0.35 &
  99.84 \small $\pm$ 0.27 &
  \multicolumn{1}{c|}{99.92 \small $\pm$ 0.13} &
  83.38 \small $\pm$ 7.70 &
  88.24 \small $\pm$ 6.15 &
  77.00 \small $\pm$ 6.68 \\
7 &
  99.52 \small $\pm$ 0.77 &
  99.85 \small $\pm$ 0.29 &
  \multicolumn{1}{c|}{99.92 \small $\pm$ 0.13} &
  89.72 \small $\pm$ 5.57 &
  89.23 \small $\pm$ 6.98 &
  78.21 \small $\pm$ 6.77 \\
8 &
  99.93 \small $\pm$ 0.13 &
  99.89 \small $\pm$ 0.15 &
  \multicolumn{1}{c|}{99.94 \small $\pm$ 0.11} &
  93.49 \small $\pm$ 4.07 &
  89.70 \small $\pm$ 6.42 &
  81.97 \small $\pm$ 6.63 \\
9 &
  99.61 \small $\pm$ 0.38 &
  99.76 \small $\pm$ 0.40 &
  \multicolumn{1}{c|}{99.91 \small $\pm$ 0.17} &
  82.48 \small $\pm$ 9.32 &
  84.85 \small $\pm$ 8.57 &
  81.37 \small $\pm$ 6.53 \\
10 &
  96.64 \small $\pm$ 2.46 &
  99.38 \small $\pm$ 0.58 &
  \multicolumn{1}{c|}{99.89 \small $\pm$ 0.15} &
  55.02 \small $\pm$ 15.42 &
  81.43 \small $\pm$ 9.77 &
  80.59 \small $\pm$ 6.83 \\
11 &
  87.04 \small $\pm$ 8.13 &
  98.40 \small $\pm$ 1.25 &
  \multicolumn{1}{c|}{99.83 \small $\pm$ 0.31} &
  29.62 \small $\pm$ 13.82 &
  82.20 \small $\pm$ 9.46 &
  80.74 \small $\pm$ 7.08 \\
12 &
  76.86 \small $\pm$ 15.23 &
  87.25 \small $\pm$ 12.39 &
  \multicolumn{1}{c|}{99.73 \small $\pm$ 0.36} &
  25.75 \small $\pm$ 26.55 &
  50.11 \small $\pm$ 32.55 &
  80.24 \small $\pm$ 6.63 \\ \bottomrule
\end{tabular}
\caption{Multilingual benchmark with additional similarity measures. The left block is with random sampling, and the right block is FAISS sampling.}
\label{tab:additional_multilingual}
\end{table*}

\begin{table*}[h]
\centering
\begin{tabular}{@{}cccccc@{}}
\toprule
Vision Model                                             &                                                       & \multicolumn{2}{c}{ViT} & \multicolumn{2}{c}{ConvNext} \\
\begin{tabular}[c]{@{}c@{}}Language\\ Model\end{tabular} &                                                       & BERT       & GPT2       & BERT          & GPT2         \\ \midrule
\multirow{3}{*}{Random}                                  & \begin{tabular}[c]{@{}c@{}}Dot\\ Product\end{tabular} & 6.32       & 6.32       & 11.39         & 8.86         \\
 & Norm            & 6.32  & 7.59  & 15.19 & 7.59  \\
 & ContraSim\_norm & 100   & 100   & 100   & 100   \\ \midrule
\multirow{3}{*}{FAISS}                                   & \begin{tabular}[c]{@{}c@{}}Dot\\ Product\end{tabular} & 5.06       & 2.53       & 7.59          & 6.32         \\
 & Norm            & 5.06  & 5.06  & 6.32  & 10.12 \\
 & ContraSim\_norm & 93.67 & 98.73 & 81.03 & 93.67 \\ \bottomrule
\end{tabular}
\caption{Image–caption benchmark results for additional similarity measures, on 4 different model pairs.}
\label{tab:additional_image_caption}
\end{table*}

Multilingual benchmark results, summarized in Table \ref{tab:additional_multilingual} show that both dot product and norm achieve better results than CKA, although achieve low results under layer prediction and image-caption benchmarks. This emphasizes the importance of multiple evaluations for similarity measures. Compared to \method{}, both methods achieve lower results. ContraSim\_norm achieves lower results compared to \method{}, under both random and FAISS sampling.

Image-caption benchmark results, summarized in Table \ref{tab:additional_image_caption}, show that under both random sampling and FAISS sampling dot product and norm achieve low accuracy. Under random sampling, ContraSim\_norm achieves perfect accuracy, while using FAISS sampling shows slight degradation compared to \method{}.

\end{document}